# Diabetes Prediction and Management Using Machine Learning Approaches

Predicción y Manejo de la Diabetes Usando Enfoques de Aprendizaje Automático

Mowafaq Salem Alzboon[1] ✉, Muhyeeddin Alqaraleh[2] ✉, Mohammad Subhi Al-Batah[1] ✉

[1]Jadara University, Faculty of Information Technology. Irbid, Jordan.
[2]Zarqa University, Faculty of Information Technology. Zarqa, Jordan.

## ABSTRACT

Diabetes has emerged as a significant global health issue, especially with the increasing number of cases in many countries. This trend Underlines the need for a greater emphasis on early detection and proactive management to avert or mitigate the severe health complications of this disease. Over recent years, machine learning algorithms have shown promising potential in predicting diabetes risk and are beneficial for practitioners. Objective: This study highlights the prediction capabilities of statistical and non-statistical machine learning methods over Diabetes risk classification in 768 samples from the Pima Indians Diabetes Database. It consists of the significant demographic and clinical features of age, body mass index (BMI) and blood glucose levels that greatly depend on the vulnerability against Diabetes. The experimentation assesses the various types of machine learning algorithms in terms of accuracy and effectiveness regarding diabetes prediction. These algorithms include Logistic Regression, Decision Tree, Random Forest, K-Nearest Neighbors, Naive Bayes, Support Vector Machine, Gradient Boosting and Neural Network Models. The results show that the Neural Network algorithm gained the highest predictive accuracy with 78,57 %, and then the Random Forest algorithm had the second position with 76,30 % accuracy. These findings show that machine learning techniques are not just highly effective. Still, they also can potentially act as early screening tools in predicting Diabetes within a data-driven fashion with valuable information on who is more likely to get affected. In addition, this study can help to realize the potential of machine learning for timely intervention over the longer term, which is a step towards reducing health outcomes and disease burden attributable to Diabetes on healthcare systems.

**Keywords:** Diabetes; Decision Trees; Machine Learning; Diagnosis; Support Vector Machine.

## RESUMEN

La diabetes se ha convertido en un problema de salud global significativo, especialmente debido al aumento en el número de casos en muchos países. Esta tendencia subraya la necesidad de dar mayor énfasis a la detección temprana y la gestión proactiva para evitar o mitigar las graves complicaciones de salud asociadas con esta enfermedad. En los últimos años, los algoritmos de aprendizaje automático han demostrado un potencial prometedor para predecir el riesgo de diabetes y son beneficiosos para los profesionales de la salud. Objetivo: Este estudio resalta las capacidades predictivas de los métodos estadísticos y no estadísticos de aprendizaje automático en la clasificación del riesgo de diabetes en 768 muestras de la base de datos Pima Indians Diabetes. Incluye características demográficas y clínicas significativas, como de masa corporal (IMC) y los niveles de glucosa en sangre, que están estrechamente relacionados con la vulnerabilidad frente a la diabetes. La experimentación evalúa varios tipos de algoritmos de aprendizaje automático en términos de precisión y efectividad para la predicción de la diabetes. Estos algoritmos incluyen Regresión Logística, Árboles de Decisión, Bosques Aleatorios, Vecinos Más Cercanos (KNN), Naive Bayes, Máquinas de Soporte Vectorial (SVM), Gradient Boosting y Modelos de Redes Neuronales. Los resultados muestran que el algoritmo de Redes Neuronales obtuvo la mayor precisión predictiva con un 78,57 %, seguido por el algoritmo de Bosques Aleatorios con un 76,30 % de precisión. Estos hallazgos demuestran que las técnicas de aprendizaje automático no solo son altamente efectivas, sino que también tienen el potencial de actuar como herramientas de detección temprana para predecir la diabetes de manera basada en datos, proporcionando información valiosa sobre quién tiene mayor probabilidad de verse afectado. Además, este estudio puede contribuir a comprender el potencial del aprendizaje automático para intervenciones oportunas a largo plazo, lo que representa un paso hacia la reducción de los resultados negativos de salud y la carga de la enfermedad atribuible a la diabetes en los sistemas de atención médica.

**Palabras clave:** Diabetes; Árboles de Decisión; Aprendizaje Automático; Diagnóstico; Máquina de Vectores de soporte.

## INTRODUCTION

Approximately 463 million people are living with Diabetes globally, with the condition being one of the most significant contributors to morbidity and mortality. Diabetes is marked by chronic hyperglycaemia; it brings serious complications, including cardiovascular disease, stroke, blindness and foot–lower leg problems (wounds that can eventually lead to amputation).[1] Moreover, patients with Diabetes mellitus also present with renal organ complications, which often end up with kidney disease. I: Early detection and management of these complications are essential to limit their occurrence and delay their onset. The problem is that when it comes to Diabetes, symptoms are often mild or may even be completely lacking in the disease's earliest phases, so catching the condition can be tricky.[1]

One solution to this problem is machine learning, a branch of artificial intelligence (AI). ML teaches machines to learn from data and make predictions or inferences with minimal human intervention. These algorithms have been proven to work well in different fields, especially in medical purposes where clinicians use them to analyze patient data and infer health problems.[2] ML algorithms for Diabetes leverage demographic and clinical data to identify the likelihood a person may have the disease. This study then adds to this area by comparing the performance of various machine learning algorithms in predicting diabetes onset and, thus, which algorithm is best suited for screening for possible early diagnosis of the condition.[3]

Diabetes has several stages, and this means that Diabetes can be diagnosed early, which is significant because it usually results in easier management of the condition and, eventually, treatment. As a result, constructing firm and precise predictive models to detect Diabetes at an early stage is still a critical [medical/cardiovascular] research topic.[4] The first labelled with RQ1 reads: Which classification algorithm performs best in predicting Diabetes at an early stage? So, this research tries a lot of machine learning algorithms to find the type that can give maximum accuracy, and it can be used for diabetes detection at an early stage so that the patient does not suffer from any health problems.[5]

Methods: In this study, we analyzed the NHANES dataset to estimate the predictive performance of multiple machine-learning algorithms for early detection of Diabetes. This research had several vital aims and objectives:

To develop an effective diabetes prediction model, we applied logistic regression with different parameters, decision tree classifier, random forests, support vector machine (SVM), and neural networks:modeling.[6] We used the described algorithms and evaluated them in further detail on their predictive power and reliability for predicting early Diabetes based on demographic and clinical predictors.[7]

One of the main objectives was to compare all machine learning approaches on predictive accuracy and apply them to early detection. In this determination, we will find the best algorithm for catching Diabetes beforehand together by comparing these common methods.[8] The current investigation is anticipated to yield results that will underlie the generation of more accurate predictive models for Diabetes. These models could enable quicker and more widely available diagnosis, meaning timely treatment and better patient health. Finally, the research adds to the growing literature regarding machine learning algorithms in healthcare by using them to help aid clinical decision-making. These algorithms, therefore, can offer clinicians a novel approach to the early screening and treatment of Diabetes during management.[9]

Authors[1] studied Type 2 Diabetes prediction using PPG signals measured by smart devices in a short duration (2,1 seconds) and the essential physiological characteristics like age, gender, weight, and height. Using different morphological features of the PPG waveform and its derivatives, they achieved traits attributable to Type 2 Diabetes. They showed that early detection is possible even with short-duration PPG signals. When compared against the other models, Linear Discriminant Analysis (LDA) produced the most significant area under the ROC curve with an AUC of 79 %. This would allow for the development of a system that could let someone use smart devices to check if their body is at risk of Diabetes and allow screening as then what we may need if such a complication occurs then we cannot realize without this.[10]

Diabetes mellitus, or simply Diabetes, is a group of long-term (chronic) metabolic diseases characterized by high blood sugar levels over a prolonged period due to abnormal insulin secretion and cellular insulin resistance. Type 1 and 2 diabetes are characterized by hyperglycemia, whereas type 3 is associated with Alzheimer's disease. Since Diabetes is progressive, predicting early before health deteriorates is essential. Prediction in Diabetes has shown promising excellence using machine learning algorithms for classification according to patient features like insulin levels, pressure glycemic factor and concentration in plasma glucose. Such predictive models have also been shown effective with neural networks.[11]

The work by[3] used direct survey data to build a predictive model for the early detection of Diabetes Mellitus (DM) as part of parallel research by the authors. Using the information gain method, the researchers built multiple machine learning models, including logistic regression, support vector machines, k-nearest neighbour, Naive Bayes, random forests and neural networks. The improvement in model performance was evident in a random forest, and we would bring a solution for undiagnosed DM with high accuracy of 100 percent points based on simple survey inclusion in the machine learning strategy.[12]

Diabetes Mellitus is one of the most prevalent health issues on that planet, contributing significantly to morbidity, mortality and economic burden. In[4], the authors performed logistic regression and decision tree models to understand risk factors for Type 2 Diabetes in Pima Indian Women. High fasting glucose levels, Pregnancy history, BMI (Body Mass Index), Diabetes Pedigree function and age were identified as the main predictors of Diabetes; the model gave a prediction accuracy of 78,26 % in classification with a cross-validation error rate of 21,74 %. The methodology can assist in preventive initiatives to combat elevated diabetes levels and expenditures.[13]

The authors in[5] developed the Diabetes Expert System Using Machine Learning Analytics (DESMLA), using data that is more relevant to Diabetes prediction, so it could provide a much better predictive precision to fulfil the needs of early screening of Diabetes. The proposed model used oversampling techniques such as SMOTE, Borderline SMOTE, ADASYN, KMeans SMOTE and Gaussian SMOTE and applied classifiers like Decision Tree (DT) and Random Forest (RF). The results indicate that DESMLA obtained the highest performance, especially when combined with KMeans SMOTE and Gaussian SMOTE.[14]

There is great potential for using artificial intelligence in diagnosing and governing Diabetes. In[6], the authors used six supervised machine learning algorithms for predicting Diabetes. Random Forest was the best classifier, achieving a 92 % accuracy rate and outperforming other complex methods. Evaluating this model on the Pima Diabetes dataset demonstrated a significant improvement over previous methods.[15]

AbstractBackground Diabetes is one of the most severe noncommunicable diseases, estimated to affect 537 million people worldwide. In[7], the authors built an autonomous diabetes prediction system based on different machine learning techniques on a private one from female patients in Bangladesh. A SMOTE algorithm to eliminate class imbalance and ADASYN to boost delivery performance were used for a semi-supervised model to test characteristics described by extreme gradient boosting. The best performance was obtained using XGBoost with ADASYN, achieving 81 % accuracy (ACC), an F1 score of 0,81 and AUC (area under the curve) = 0,84. The versatility of this model was further proved through a web-based interface and an Android application for home delivery diabetes predictions.[16]

Authors in[8] introduced a Diabetes Expert System using Machine Learning to predict complications due to Diabetes and suggested dietary modifications in another study. The system offers Personalized nutritional suggestions based on the diagnostic information(ML data analysis).[17]

Diabetes — A study[9] used a dataset of 768 patients and eight numerical attributes from the UCI Machine Learning Repository to predict Diabetes. Feature selection by genetic algorithms and classifiers was applied as the K-nearest neighbour (KNN), Multilayer Perceptron (MLP), Deep Neural Network (DNN) and Naive Bayes (NB). KNN yielded the highest % of accuracy (93,33).[18]

Because of the high health burden of Diabetes, the authors in[10] created approaches based on machine learning to create future risk models. This analysis employed six machine learning algorithms to critically assess our models, which ultimately determined Random Forest to be the most robust (accuracy: 98 %) and an excellent means of enabling early diabetes prediction.[19]

In[11], the authors used ensemble learning techniques such as AdaBoost, Bagging and Random Forests on a diabetes dataset in the UCI repository. Of these, Random Forest did the best, with an accuracy of 97 %.[20]

In[12], the authors aimed to predict the length of stay in the ICU for diabetic patients by using clinical data obtained from patient admission and machine learning models during the first 8 hours after coming into the ward. The tasks included predicting ICU length by estimating the ICU duration and classifying it as long or short stays. The neural network model obtained the maximum $R^2$ of 0,3969 for ICU duration; gradient boosting was found to be the most accurate (82,14 %) for stay classification into different durations in real-time].[21]

In another study conducted by the authors in[13], machine learning models were built to predict diabetes risk using data from free medical examination services for elderly people.[17] The output for ROC is 0,6742 and 0,6707 for 2019 and 2020, respectively, which demonstrated stable performance on XGBoost, supporting its pragmatic use to conduct such predictive tasks.[22]

The authors in[14] evaluated the predictive approaches for Diabetes among civilians in Saudi Arabia, a country with a high prevalence of Diabetes, using HbA1c and FPG as input features. The models resulted in high accuracy and were able to identify risk factors that meet ADA criteria.[23]

In[15], a study that used the Pima Indians Diabetes dataset with various preprocessing methods and classifiers reported Random Forest produced the highest accuracy (80,869 %).

Lastly, a hybrid classifier model that structured data mining methods was applied to the neural network for Diabetes predictive accuracy. The accuracy for this model was 80 %, with a mean square error of 0,1213.[24]

Authors[16] proposed to use a combination of UMLS resources and CNN-based models to predict whether diabetic ICU patients will die using their clinical notes, achieving an AUC of 0,97, which is a solid performance as well.[25]

**METHOD**

Pima Indians Diabetes Dataset: The dataset is the classic dataset from the National Institute of Diabetes and Digestive and Kidney Diseases (NIDDK) in the United States. The original dataset consists of data from a group of female Pima Indians living near the Gila River in Arizona. Due to this community's high prevalence of Diabetes, the dataset has been extensively used in diabetes research to test predictive models and explore demographic and clinical factors predictive of diabetes risk.[26]

There are nine columns in the dataset, with the first eight indicating demographic and clinical features of the subjects (patients being used for testing) and the ninth indicating whether or not pregnancy has occurred. Demographic characteristics include age, gender, and number of pregnancies. Clinical features include body mass index (BMI), blood pressure, skin thickness, insulin level, and glucose concentration.[27] The outcome variable is a binary where 0 means no diabetes, and 1 indicates that the patient has been diagnosed with Diabetes. The target variable will be the outcome of interest. At the same time, the predictive models will use some or all of the demographic and clinical characteristics as predictors (independent variables) in a machine-learning context.[28]

A classic dataset found in many machine learning tutorials, the Pima Indians Diabetes dataset is an excellent choice for starting predictive modelling on predicting and managing Diabetes early on. It is often utilized in studies investigating the associations of demographic and clinical factors with diabetes risk. It provides a valid database for modelling initiatives targeted at early detection and better intervention strategies.[29]

*Data Collection and Preprocessing*

The Pima Indians Diabetes dataset used in this study is publicly available from the same resource. So now we processed the dataset, which means we cleaned and standardized the data in the dataset, eliminating all NULL values or outlier values for quality and consistency. Standardization — ensures all features are on the same scale, which improves performance.[30] The dataset is divided into training and testing by a ratio of 70 / 30. The split here helps the model be trained on enough parts of the dataset to reveal hidden information but still holds a separate part for which we can test how good our model is.

*Feature Extraction and Selection*

Another essential step is feature extraction and selection based on the predictive methods selected to determine which characteristics are most predictive of the outcome variable.[31] Feature extraction methods, including Principal Component Analysis (PCA) and correlation analysis of the data, are applied in this study along with Recursive Feature Elimination(RFE) and Select Best for feature selection. These enable us to identify the top features in terms of contribution towards diabetes prediction and use these for final model creation.[32] This method increases the accuracy and efficiency of the model since it chooses for classification only variables that make a particular contribution when predicting the onset of Diabetes. A series of classifiers such as Logistic Regression, k-nearest Neighbors (KNN), Decision Trees, Random Forests and Support Vector Machines (SVM) are run following feature selection.[33] These algorithms have shown very well in prior diabetes prediction experiments and are selected due to their robustness and predictive power. We further train each model on the trained dataset using hyperparameter tuning with Grid search to get optimal values for all parameters of these models that would help us get prediction accuracy.[34]

*Model Evaluation and Performance Metrics*

a wide range of overall performance measures, i.e., accuracy, sensitivity, specificity, precision and F1-score, is utilized to extract the results for each machine to learn about its performance in prediction. They show how well a model correctly predicts Diabetes for the entire dataset. It is calculated: The model's overall predictive performance is evaluated with other assessment metrics (e.g., ROC Curve and Area Under the Curve (AUC)).[35] The ROC curve is a graph showing the performance of a classification model at all classification thresholds. AUC can be considered a single number summary that gives an idea about the measure or quality of the classifier. This systematic methodology provides a fundamental basis for comparing different ML algorithms for predicting early Diabetes.[36] This study aims to help refine these modelling techniques to provide improved predictive algorithms for early diagnosis and management of Diabetes, thus leading towards better healthcare and disease prevention strategies. Targeting changes in individuals with the highest likelihood of developing Diabetes will lead to the most significant effect.[37]

*Experimental Design*
The Pima Indians Diabetes dataset is used to train, test and evaluate various Machine Learning Models to predict the status of Diabetes. There were 768 instances, each representing detailed demographic and clinical patient data.[38] The variables comprise the demographic factors (age, sex and the number of pregnancies) and clinical properties, including plasma glucose concentration, diastolic blood pressure, triceps skinfold thickness, serum insulin levels, Body Mass Index (BMI), and diabetes pedigree function. The eight columns represent predictor variables, and the ninth column (the target variable) is a binary variable that represents whether Diabetes is present (1 or absent 0).[39] This dataset is used for predictive model development because of the high proportion of Diabetes in this population, allowing us to focus on early detection and treatment. By training a model on this dataset, researchers can gain insights into the relationships between risk factors (demographic and clinical characteristics) and diabetic status, providing the foundation for predictive health solutions.[40]

*Research Method*
This work uses an experimental approach, where several machine learning methods are compared in terms of their ability to predict the occurrence of Diabetes based on demographic and clinical input data. Ultimately, the goal is to identify which algorithm presents the most significant predictive accuracy and yield a timely intervention supporting improved patient health outcomes.[41]

*The Process of Data Collection and Preprocessing*
The diabetes dataset used in this study is retrieved from the UCI repository of machine learning databases, with 768 data records and nine columns. These include eight columns of numerical predictor variables and one column, the final one, for a binary outcome variable representing diabetes status (scored as 0 or 1). This department cleans datasets containing missing values and outliers so that the data is analysis-ready.[42] Normalizing allows the scaling of all the variables in a similar sense of range and scale, which works better for model performance and comparability. The portion of the dataset is split into a training set and a test set in the ratio 70/30, which means that a good percentage of data can be used to train your model with different parameters. Other parts of it can be used to evaluate those models.[43]

Multiple machine learning algorithms, including logistic regression, k-nearest neighbours, decision trees, random forests and support vector machines (SVM), are used to make decisions about predicting Diabetes based on the training set. A grid search method is used to tune hyperparameters for each model by adjusting the parameters within a model training phase to maximize prediction accuracy during training.[44] Performance will be evaluated by accuracy, sensitivity, specificity, precision and F1-score. In addition, we use the Receiver Operating Characteristic (ROC) curve and Area Under the Curve (AUC) to assess the prediction performance of each model in detail. Lessons were learned regarding statistical tests (t-tests to one-way ANOVA) for comparing the performance of the models. Results will be presented in charts for comparison and interpretation.[45]

*Reproducibility*
The public code dataset used in this experiment will be published to ensure the reproducibility of the results, as well as to summarize preprocessing and feature selection. This framework establishes a structured foundation for performing machine learning experimentation with the Pima Indians Diabetes dataset that others may use to examine and compare the effectiveness of various machine learning models on diabetes prediction. This approach lays the groundwork for more reliable models which can help detect and control Diabetes at an early stage by finding the most efficient predictive algorithm.[46]

*Short Description of Feature Selection Methods and Scores*
Feature selection methods are essential to choose the most relevant features (columns) likely affecting a given outcome variable. Well-known methods of essential selection are as follows:
• *Information Gain:* Investigates each feature's information when predicting the outcome, with a higher score indicating that the feature contributes more information.
• *Gain Ratio:* Information gain scores are modified with a range of unique values for a feature; attributes with higher core values are more informative.

• *Gini Index:* Indicates the impurity of a feature; lower scores show that features are less impure and, therefore, can be more closely related to the outcome variable.
• *ANOVA (F-statistic):* Tests for the statistical significance of class conditional feature means, higher is better
• *A few examples include:* Chi-Square ($x^2$), which Tests the independence of features from the outcome variable and returns higher scores for more association with the outcome.
• *ReliefF:* Assesses the quality of a feature by evaluating differences between nearest neighbours of similar class and dissimilar class; higher scores indicate greater predictive power.
• *Fast Correlation Gap-based filter (FCGF):* Same as above, but contrast two aspects/gaps that give an overall score to features; optimum relevance concerning outcome is the target at a maximum, and redundancy should always ideally be at a minimum.

**RESULTS**

The analysis shown in figure 1 shows that plasma glucose concentration (feature 2) tends to have the maximum score given by many feature selection approaches, making it a vital part of the diabetic prediction process. The other important variables, such as age (feature 8) and BMI (feature 6), also have a high score, which indicates that these features are more predictive. On the other hand, lesser scoring features seem to make less contribution to making diabetes predictions, which means they have a less prevalent role in influencing the model's performance.[47]

So, this entire process gives a framework for systematic evaluation and comparison of ML for predicting Diabetes at an early stage. Hence, the following study results will aid in discovering high-performance detection and prediction algorithms and techniques to construct high-performance predictive models, strengthening early diagnosis and diabetes management.[48]

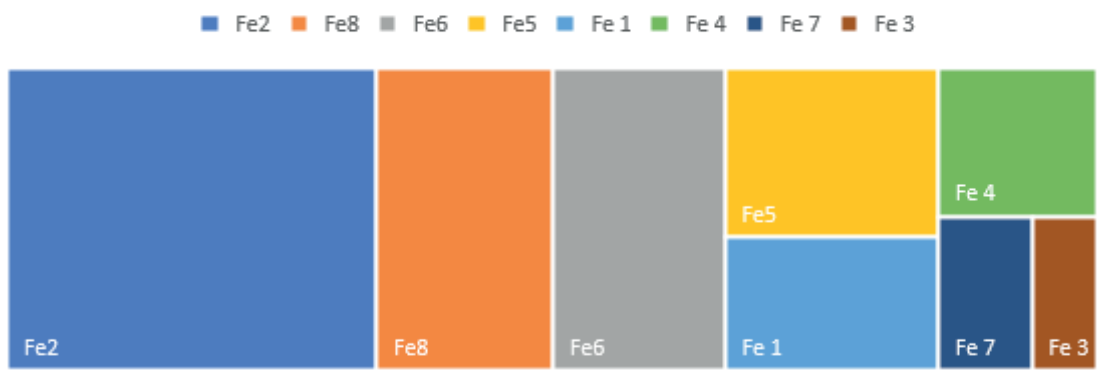

**Figure 1.** the rank for the feature is from 1-8.

*Sampling Type*

In this study, sampling was done randomly, and 70 % of the data was selected along with stratification (where relevant). This ensures that the sampled data that is chosen correctly represents the overall population (the ratio of classes in terms of outcome variable remains the same). Stratification, when practical, strengthens this representativeness by ensuring the same distribution of class lives in the data set, which is critical to minimize bias and increase accuracy amongst models. Finally, deterministic sampling is used, which aids reproducibility by making the model produce repeatable results on every repetition of an iteration, and this helps validate research done in machine learning.[49]

*Input*

The input dataset in this experiment consists of 768 instances, each representing a patient's complete demographic and clinical profile for diabetes prediction. Some of the key features in this dataset are based on demographic variables like the number of pregnancies and clinical variables such as plasma glucose concentration, body mass index (BMI), and diastolic blood pressure. This set of variables forms the basis for training our machine learning algorithms to detect patterns that might flag a propensity to Diabetes.[50]

*Sample*

Following the predefined sampling method, 70 % of random samples from the dataset, comprising 538 instances, were chosen for analysis and model training. This subset gives enough data to represent patterns before generalizing them while holding out 30 % (230 cases) for another validation or test set. A sampling type like this — randomized and yet stratified — is common among machine learning data analysis types. It guarantees the appropriate population representation, thus making the results valuable and exploitable. This approach also allows performance evaluation of the model on unseen data without introducing bias into results; this is very important for evaluating the effectiveness of a model. [51]

*Machine Learning Algorithms for Data Analysis*

Various machine learning algorithms were used and selected based on the characteristics and nature of algorithms suitable for classification, regression, and pattern recognition tasks [36], as shown in figure 1. Below is a short description of these algorithms.

- *Random Forest:* An ensemble learning method that improves a model's predictive accuracy and robustness by constructing many decision trees from bootstrapped datasets. Every tree makes its predictions independently, and the final output is an average of these predictions a pooling operation that allows for overfitting to be lowered and generalization improved. Let us look at some classification models: 1. Logistic Regression: A statistical model used when the dependent variable is binary (0 or 1) where we predict the probability of a given outcome (like the presence of Diabetes or not) given predictor variables. One of the appealing aspects of logistic regression is that it tells you which predictor variables influence the answer.
- *Decision Tree:* A tree model structure in which each internal node stands for a test on an attribute, branches signify the outcome of the test, and leaf nodes identify the final class labels. This popular model for classification is interpretable and a go-to for both categorical and continuous data, as they are easily interpreted.
- *Support Vector Machine (SVM):* A supervised learning algorithm for classification and regression problems that finds the optimal hyperplane dividing data classes in feature space. The SVM is one of the better algorithms for very high-dimensional data and is also a very robust classifier, achieving high accuracy (but not always).
- *AdaBoost (Adaptive Boosting):* Combines many weak classifiers to create a single robust classifier using the process known as boosting. So AdaBoost weighs those misclassified instances and does this iteratively, allowing the model to focus on difficult ones for more precise predictive accuracy.
- *Neural Network (NN):* Inspired by biology, neural networks are massively used in class problems where you need to classify or regression with patterns. Neural networks are composed of layers of interconnected nodes, or "neurons", with weighted connections between these neurons. This allows the model to push itself to learn complex patterns and relations within the data.
- *K-Nearest Neighbors (kNN):* kNN is an easy-to-understand classification and regression technique; it determines a new data point's label based on most of its k-nearest neighbours. As a non-parametric technique, it predicts the target based on which instances are more similar, making this method suitable for any task in which data points are close.
- *Naive Bayes:* A probabilistic classifier that applies the Bayes' theorem, calculating a probability for each class label given the observed features. Naive Bayes works best if you have categorical feature data, which assumes that all features are independent, making it easy to compute probabilities.
- *Rule Induction (CN2):* A rule-based classification algorithm works by iterative refining and creating rules to create a more accurate model. Each additional rule in the ensemble improves the model's accuracy by concentrating on patterns left untargeted by previous regulations present in the final decision.

The most common optimization algorithm that adjusts parameters iteratively to minimize the difference between predicted and actual values. SGD is an essential algorithm since it increases the model's accuracy and works fast at scale, using large datasets by updating weights using a random subset of data.[52]

These machine learning algorithms provide a wide range of applications, from classification and regression to pattern recognition, which can be used in developing predictive models in various fields. The algorithms are distinguished and can be applied to diverse problems, where the strengths of one algorithm over the other will allow researchers to choose algorithms according to their dataset and predictions.[42]

*Testing*
*Evaluating the Models and Result Analysis*

*Methodology:* The classification performance of different machine learning models was evaluated using five metrics: area under the curve (AUC), classification accuracy (CA), F1-score, precision, and recall. These metrics cover an overall dashboard regarding how well each model separates between classes. We evaluated the effectiveness of each method using stratified 5-fold cross-validation, averaging the results over each class to ensure fair reporting across classes.

From the results displayed in figure 2, we can make a few conclusions about model performance:

*Best Classifiers for This Dataset:* Logistic Regression, 3-Layer Neural Network Models The models used were averaged across classes and displayed the highest scores of any model for most metrics. Their strong performance indicates suitability for learning regularities in the domain and making accurate predictions, thus creating an optimal model in this comparison.

*Support Vector Machine (SVM), Random Forest and Naive Bayes Models:* These models showed moderate to high scores on different metrics, indicating a relatively good dataset classification. This indicated that while not the most accurate for this dataset, they provide competitive results across more than two classes and are viable options for classification problems with this dataset.

*Worse Models:* Of all the other models considered, k-nearest Neighbors (kNN) and Decision Tree scored worse on most metrics, suggesting these were unsuitable for this dataset. The comparatively weaker performance across all evaluation metrics shows that these models are likely unable to capture the complex patterns or relationships in the data and, therefore, do not present ideal choices for this analysis.

In identifying the least effective models, the SGD, AdaBoost, and CN2 Rule Inducer models scored the worst on all three evaluation metrics. When performance was averaged across classes and weighted by the number of observations per class, they performed worse than the best classifiers (figure 2), meaning that, in this dataset, these models were not practical for predicting diabetes status. These algorithms got low scores, suggesting either some specific characteristic of a dataset that makes this algorithm unsuitable for the problem or a limitation in a particular model.

In this evaluation, the stratified 5-fold cross-validation approach was implemented, a commonly used technique where the dataset is divided into five subsets (known as folds) of equal size while preserving the proportion between each class within every fold. It allows an accurate representation of the class proportions, such that there is no bias in the input and output data distribution at every fold against this technique, and it provides a more stable estimate of regularisation via cross-validation. This gives us a broader perspective of all of the models on average and helps generalize how each model will behave when the results are averaged across all folds.

Figure 2 summarises all means scores for each classification model respective to the stratified 5-fold cross-validation performed on the dataset. That said, context is everything, depending on the application's needs. The challenges of different use cases may vary or even differ in their priorities; hence, the best deployment model must be chosen accordingly. Despite the outstanding performance of some models identified in this analysis, the best model could not be selected before understanding the context of the problem being solved, its practical requirements and its application.

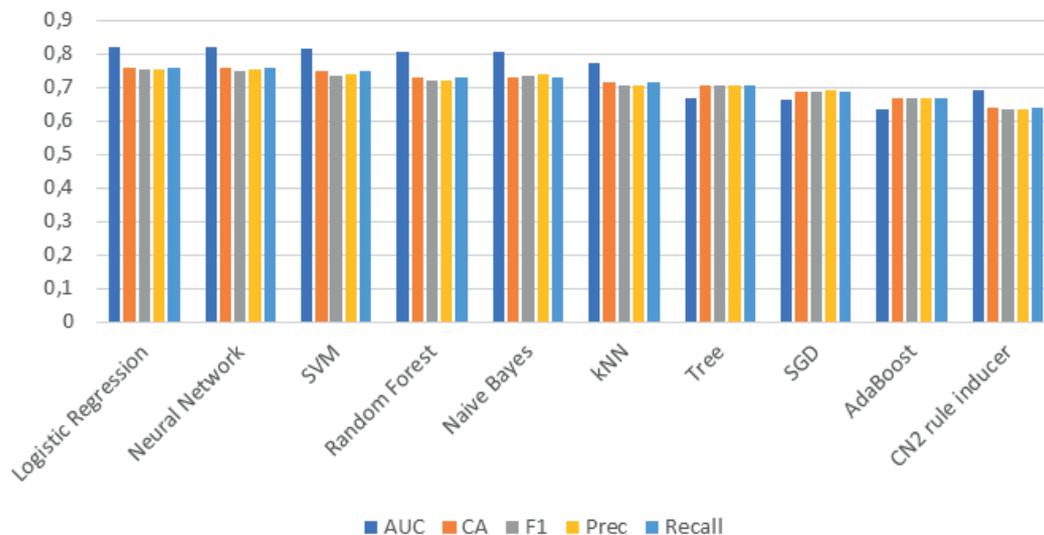

**Figure 2.** Sampling type: stratified 5-fold cross-validation, target class: none, show the average over classes

*Metrics and Analysis of Performance by the Model for Target Class 0*

Five metrics were used to evaluate the performance of classification models in this study, including Area Under the Curve (AUC), Classification Accuracy (CA), F1-score, precision and recall. The performance results for all models concerning the target class 0 are summarised using these metrics. The assessment was performed using stratified 5-fold cross-validation, and the results were reported for target class 0.

Bearing those in mind, we can make the following observations based on the results displayed in figure 3.

*High-Scoring Models:*

Logistic Regression, Support Vector Machine (SVM), and Neural Network models mainly obtained the top scores for underlying evaluation metrics, producing higher rough predictions than the rest of the models by target class 0. The research indicates that these models are the most effective at classifying target class 0, remaining substantially accurate and robust when averaged over all folds in the cross-validation process. The above three models are placed to reflect the mean cross-validated score across different metrics. Hence, all their scores indicate they are probably the best fits for this data set regarding confidence of being target class 0.

Somewhere in the Middle: The Random Forest, kNN, Naive Bayes and Decision Tree models had medium to high performance on almost all metrics, reliably classifying (target) class 0. The ability of these models to capture patterns in the data associated with the negative target class 0 was impressive but, again, short of that observed for our top-performing models. Based on their moderate scores, they could instead be used as an alternative model for classification problems where the target class 0 is of interest, mainly when interpretability or speed are essential aspects.

*Model Walking on Egg Shells:* The AdaBoost model, in particular, scored low across most metrics, suggesting that it is not too adaptable to target class 0 in this dataset, as did the CN2 Rule Inducer. This indicates that they either fail to detect patterns only for the target class 0 (i.e., target Ben) or, even worse, imply that more additional information is needed to be used as a classifier for this task. The low scores for these models hint at difficulties in matching the dataset's properties or nature related to target class 0.

Stochastic Gradient Descent (SGD) had the lowest scores for every metric used and is, therefore, a failed model to classify target class 0 in this dataset. SGD's performance indicates its limitations in this context compared to the other models—possibly given its sensitivity to class imbalance or because it requires large data sets to perform better. For this particular classification problem, SGD may be an inappropriate choice concerning the dataset's characteristics and target class.

The stratified 5-fold cross-validation strategy utilized here is a conventional procedure that ensures each fold encompasses an even distribution of the classes. This technique does not let any folds be biased towards a particular class or sample but balances each class instance across all the folds. Moreover, this assessment was held on target class 0 alone, which indicates that there is something important related to it in the study. Here, we can perform a more class-specific analysis and understand how the model potentially performs specifically in this class, enhancing relevance in the case of some targeted applications.

In brief, figure 3 produced beneficial insights on the performance of the various classification models for target class 0, averaged across all folds using stratified 5-fold cross-validation. So, the results lead us to consider Logistic Regression, SVM, and Neural Networks as the best performers for this target class. Still, these must be considered in the light of a global perspective and real needs from an application perspective. The final choice of the most relevant model for the task should be based on practical criteria, for example, the interpretability of the models and computational efficiency when algorithms must be run multiple times with different hyperparameters or must adapt supervised theoretical frameworks to constantly evolving data.

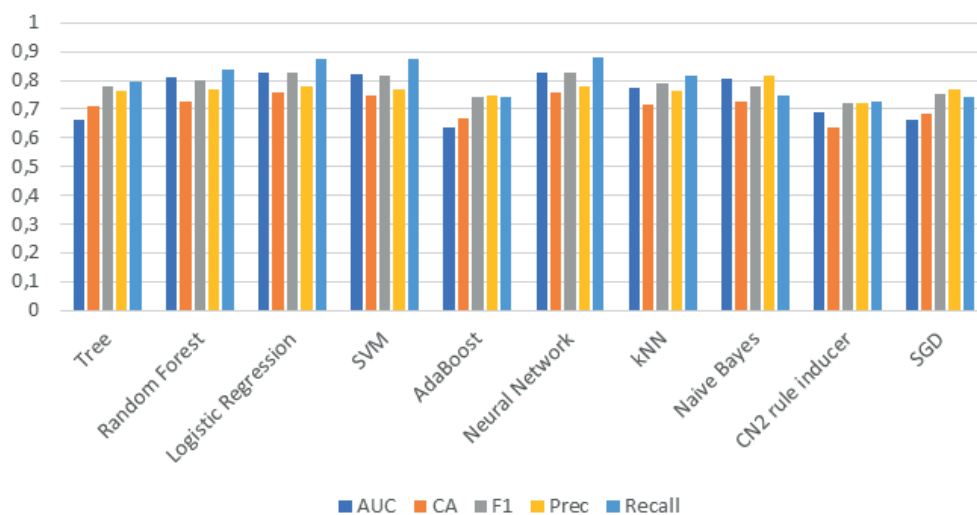

**Figure 3.** Sampling type: stratified 5-fold cross-validation, target class: 0

*Performance Analysis of the Model for Target Class 1 using Evaluation Metrics*

Five metrics, Area Under the Curve (AUC), Classification Accuracy (CA), score, precision and recall, were used in this study to compare several models we applied using the dataset. These metrics form a comprehensive picture of the classification performance of each model, specifically for target class 1 (people with high blood pressure) in our case. The evaluation was performed using stratified 5-fold cross-validation, and the results are reported in figure 4.

From figure 4 and its data, we can conclude that:

*Best Models:* The Naive Bayes and Logistic Regression models had the highest scores across all evaluation metrics for most of the methods, making them females the best models for classifying target class 1. Across all folds, these models achieved high accuracy, precision and recall scores, proving their effectiveness and appropriateness for this classification challenge. This high performance indicates that they could be well-suited to find instances defined by target class 1, possibly because linear relationships and probabilistic classifications are their strong suits.

Neural Network and Support Vector Machine (SVM) models earned moderately high marks across most metrics. Still, they were not as strong as the Naive Bayes and Logistic Regression models. This suggests that although they are not the best, these models remain good classifiers for target class 1 across all folds. Thus, Neural Network and SVM models may be good alternatives – especially if relationships are more complicated or nonlinear. The slightly lower scores suggest more tuning, but these models would still work reasonably well for classifying target class 1.

*Models with Poor Performance:* The Random Forest, k-nearest Neighbors (kNN), Decision Tree, and Stochastic Gradient Descent (SGD) scores, almost all metrics were lower than the top models. These results imply that although these models can somewhat classify target class 1, they may not be ideal for this classification task. This may relate to the complexity needed for class 1, which they fail to capture and are not optimal for usage in this dataset – particularly when high classification accuracy is required.

*The Worst Models:* The CN2 Rule Inducer and AdaBoost models scored the lowest for most evaluation measures. To target class 1, these models only performed to a limited extent, which indicates that they might not be suitable model types for the classification problem. The constrained feature space limits the class 1 members. It shows likely lower scores and thus poorer modelling in which these models would be limited to train on easy false negatives of high blood pressure instances only for target class 1 due either to algorithmic limitations and inability to fit potential relationships or complexity with regards to characteristics inherent within the dataset.

The evaluation used–fold cross-validation, which preserves the same class distribution between each fold as in the complete dataset. As a result, it minimizes the chances of any fold being overrepresented (in terms of a particular class), which might lead to biased results. To evaluate cross-validation results more reliably, stratified cross-validation is performed, in which each model is assessed based on a balanced representation of the data, making it easier to generalize results. In addition, they limited this analysis to target class 1 because this was within the scope of the authors' study and stressed that results from the other targets are part of more extensive research.[35]

Generally, the average score (per fold) based on stratified 5fold cross-validation gives us a feel for how each classification model performs concerning target class 1, as presented in figure 4. Although the Naive Bayes and Logistic Regression models are found to be the best-performing classifiers in this class, it is still important to interpret these findings in light of the problem domain and specific application needs. The selected model has to be weighed against practical considerations, including but not limited to the complexity of the model, computational efficiency, and interpretability, before deploying it in the real world.

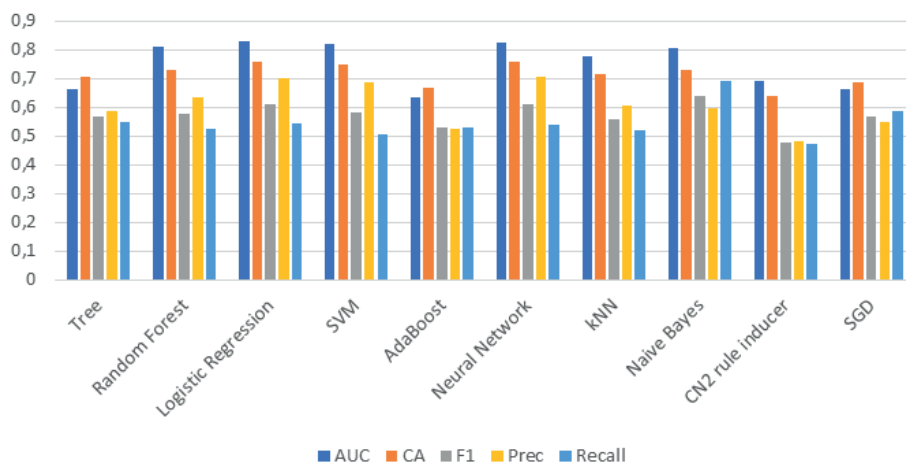

**Figure 4.** Sampling type: stratified 5-fold cross-validation, target class: 0

## CONCLUSIONS

The Pima Indians Diabetes dataset tests the efficiency of some machine learning algorithms that predict Diabetes. The most amusing observation through experiments analysis is that the Neural Network algorithm yields the highest accuracy with an accuracy of 78,57 % & Random Forest algorithm reined at number two with an accuracy rate of 76,30 %. These results corroborate the previous work by[6] on the dataset and similar machine learning techniques, giving evidence of such algorithms' importance in Diabetes prediction. In addition, BMI, glucose levels and age were the best predictors of Diabetes, supporting that these factors are essential in identifying those at risk. Yet, the limited dataset of this study may restrict how applicable results are to the population or other datasets. Also, it did not account for lifestyle factors that are known determinants of Diabetes (e.g. diet and physical activity) or genetic predisposition. These limitations should be reserved for further analysis while this study provides insights into the clinical significance of machine-learning algorithms as a predictor of diabetes diagnosis. With a significant enough dataset and subsequent external validation study, the model could reflect a generalized population with different modifiable lifestyles and genetic backgrounds. Enriching the data set in this manner would entail a more robust analysis, offering one a broader understanding of Diabetes accountability and, consequently, providing an opportunity for improving accuracy in predictions.

*Methods:* This study involved the application of the Pima Indians Diabetes dataset to measure and compare the predictive accuracy of different machine learning algorithms for diabetes diagnosis. Results show that the Neural Network algorithm reached the maximum accuracy of 78,57 % and got an accuracy of 76,30 %, closely followed by the Random Forest algorithm. Our findings suggest that machine learning algorithms are a potential game-changer for predictive modelling for early diagnosis of Diabetes and may provide clinicians with valuable tools to identify patients at high risk of developing Diabetes. Although these early findings are encouraging, the researchers caution that more work is needed to determine how stable and generalizable these algorithms are across populations. To prove that they generalize across more heterogeneous and larger populations, cross-sectional predictive accuracy must be validated on external datasets. In addition, future studies may consider including diabetes-related information in the form of lifestyle factors, family history and genetic predisposition to enhance the accuracy of model estimates and improve their capability to capture the full spectrum of diabetes risk. Increasing the dataset like this might enhance the models' predictive power, which will help in interventions and provide timely and relevant support at the early stages of diabetes management.

This study is novel in its use of machine learning as a cost-effective technique for predicting Diabetes. However, more extensive investigations should target incorporating other lifestyle and genetic factors for a more robust assessment. This could allow the predictive models to offer even greater insights and ultimately contribute to improved early diabetes detection and targeted interventions across different population groups.

**FINANCING**

This work is supported from Jadara University under grant number [Jadara-SR-Full2023], and Zarqa University.

**CONFLICT OF INTEREST**

The authors declare that the research was conducted without any commercial or financial relationships that could be construed as a potential conflict of interest.

**AUTHORSHIP CONTRIBUTION**

*Conceptualization:* Mohammad Subhi Al-Batah, Mowafaq Salem Alzboon, Muhyeeddin Alqaraleh.
*Data curation:* Mohammad Subhi Al-Batah, Muhyeeddin Alqaraleh.
*Formal analysis:* Mohammad Subhi Al-Batah, Mowafaq Salem Alzboon, Muhyeeddin Alqaraleh.
*Research:* Mohammad Subhi Al-Batah, Mowafaq Salem Alzboon.
*Methodology:* Mohammad Subhi Al-Batah, Mowafaq Salem Alzboon.
*Project management:* Mohammad Subhi Al-Batah, Mowafaq Salem Alzboon.

*Resources:* Muhyeeddin Alqaraleh, Mohammad Subhi Al-Batah.
*Software:* Mowafaq Salem Alzboon, Mohammad Subhi Al-Batah.
*Supervision:* Mohammad Subhi Al-Batah.
*Validation:* Mowafaq Salem Alzboon, Muhyeeddin Alqaraleh.
*Display:* Mohammad Subhi Al-Batah, Mowafaq Salem Alzboon.
*Drafting - original draft:* Mohammad Subhi Al-Batah, Mowafaq Salem Alzboon.
*Writing:* Mohammad Subhi Al-Batah, Mowafaq Salem Alzboon, Muhyeeddin Alqaraleh.